%% file: main.tex
\documentclass[runningheads]{llncs}

\usepackage{eccv}

\usepackage{eccvabbrv}

\usepackage{graphicx}
\usepackage{booktabs}

\usepackage[accsupp]{axessibility}  % Improves PDF readability for those with disabilities.

\usepackage[pagebackref]{hyperref}

\usepackage{orcidlink}
\usepackage{multirow}
\usepackage[table]{xcolor}
\usepackage[most]{tcolorbox}
\begin{document}

% ---------------------------------------------------------------
% TODO REVIEW: Replace with your title
\title{QMoP: Query Guided Mixture-of-Projector for Efficient Visual Token Compression} 

% TODO REVIEW: If the paper title is too long for the running head, you can set
% an abbreviated paper title here. If not, comment out.
\titlerunning{QMoP}

% TODO FINAL: Replace with your author list. 
% Include the authors' OCRID for the camera-ready version, if at all possible.
\author{Zhongyang Li\inst{1} \and
Yaqian Li\inst{2} \and 
Faming Fang\inst{1}\textsuperscript{*}\and 
Rinyoichi Takezoe\inst{2} \and 
Zi-Hao Bo\inst{2} \and 
Cheng Qian\inst{2} \and 
Mo Guang\inst{2} \and 
Guixu Zhang\inst{1} \and
Kaiwen Long\inst{2}\textsuperscript{*}
}

% TODO FINAL: Replace with an abbreviated list of authors.
% \authorrunning{F.~Author et al.}
% First names are abbreviated in the running head.
% If there are more than two authors, 'et al.' is used.

% TODO FINAL: Replace with your institution list.
\institute{East China Normal University,  \and
Li Auto Inc.}

\maketitle
\begingroup\renewcommand{\thefootnote}{*}\footnotetext{Corresponding authors: Faming Fang and Kaiwen Long.}\endgroup

\input{sec/0_abstract}
\input{sec/1_intro}

\input{sec/2_related}

\input{sec/3_method}

\input{sec/4_experiment}
\input{sec/5_conclusion}

% \clearpage  % TODO FINAL: This \clearpage needs to be removed from both review and camera-ready versions.

% \section*{Acknowledgements}
% Please insert your acknowledgments here.

% ---- Bibliography ----
%
% BibTeX users should specify bibliography style 'splncs04'.
% References will then be sorted and formatted in the correct style.
%
\bibliographystyle{splncs04}
\bibliography{main}
\input{supp_for_main}
\end{document}

%% file: sec/0_abstract.tex
\begin{abstract}
Multimodal large language models suffer from severe computational and memory bottlenecks, as the number of visual tokens far exceeds that of textual tokens. While recent methods employ projector modules to align and compress visual tokens into text-aligned features, they typically depend on fixed heuristics that limit adaptability across diverse scenarios.
In this paper, we first propose \textbf{Query Guided Mixture-of-Projector} (\textbf{QMoP}), a novel and flexible framework that adaptively compresses visual tokens via three collaborative branches: (1) a pooling-based branch for coarse-grained global semantics, (2) a resampler branch for extracting high-level semantic representations, and (3) a pruning-based branch for fine-grained token selection to preserve critical visual detail.
To adaptively coordinate these branches, we introduce the \textbf {Query Guided Router} (\textbf{QGR}), which dynamically selects and weights the outputs from different branches based on both visual input and textual queries. A Mixture-of-Experts-style fusion mechanism is designed to aggregate the outputs, harnessing the strengths of each strategy while suppressing noise.
To systematically evaluate the effects of Visual Token Compression, we also develop \textbf{VTCBench}, a dedicated benchmark for evaluating the information loss induced by visual token compression.
Extensive experiments demonstrate that despite relying on fundamental compression modules, QMoP outperforms strong baselines and delivers significant savings in memory, computation, and inference time.
\keywords{Vision-Language Understanding
 \and Multi-modal LLM \and Token Compression}
\end{abstract}

%% file: sec/1_intro.tex
\section{Introduction}
% 多模态大语言模型（MLLMs）（cite{llava, llava15, Minigpt-4}）最近通过将视觉编码器（cite{clip, SAM, dinov2, siglip}）与预先训练好的大语言模型（LLMs）（cite{gpt4, qwen, llama, vicuna}）整合在一起，在视觉语言任务中展现出了令人印象深刻的能力。在这个框架中，视觉编码器将图像转换为视觉标记，然后将其投射到 LLM 的文本嵌入空间，并与文本标记一起处理，以完成推理和生成任务 （cite{vqav2,sqa,vizwiz, mmvet}）。然而视觉标记往往远远超过文本标记,直接将所有视觉标记传递给 LLM 会导致内存占用和计算量过大。
% ，由于视觉输入的信息密度比文本低，在空间和时间存在（cite{10tokenmerge,11llavamerge}），
Multimodal Large Language Models (MLLMs) \cite{llava, llava15, Minigpt-4} have recently shown impressive capabilities in visual-language tasks by integrating vision encoders \cite{clip, siglip} with pre-trained Large Language Models (LLMs) \cite{gpt4, qwen}. In this framework, vision encoders convert images into visual tokens, which are then projected into the LLM’s text embedding space and processed together with text tokens for reasoning and generation tasks \cite{vqav2, sqa}. Early approaches employ simple multi-layer perceptrons (MLPs) as projectors, which keep the number of tokens unchanged and only transform visual tokens along the channel dimension into the text embedding space. However, the large number of visual tokens limits further development.

\begin{figure*}[t!]
  \centering
  \includegraphics[width=\linewidth]{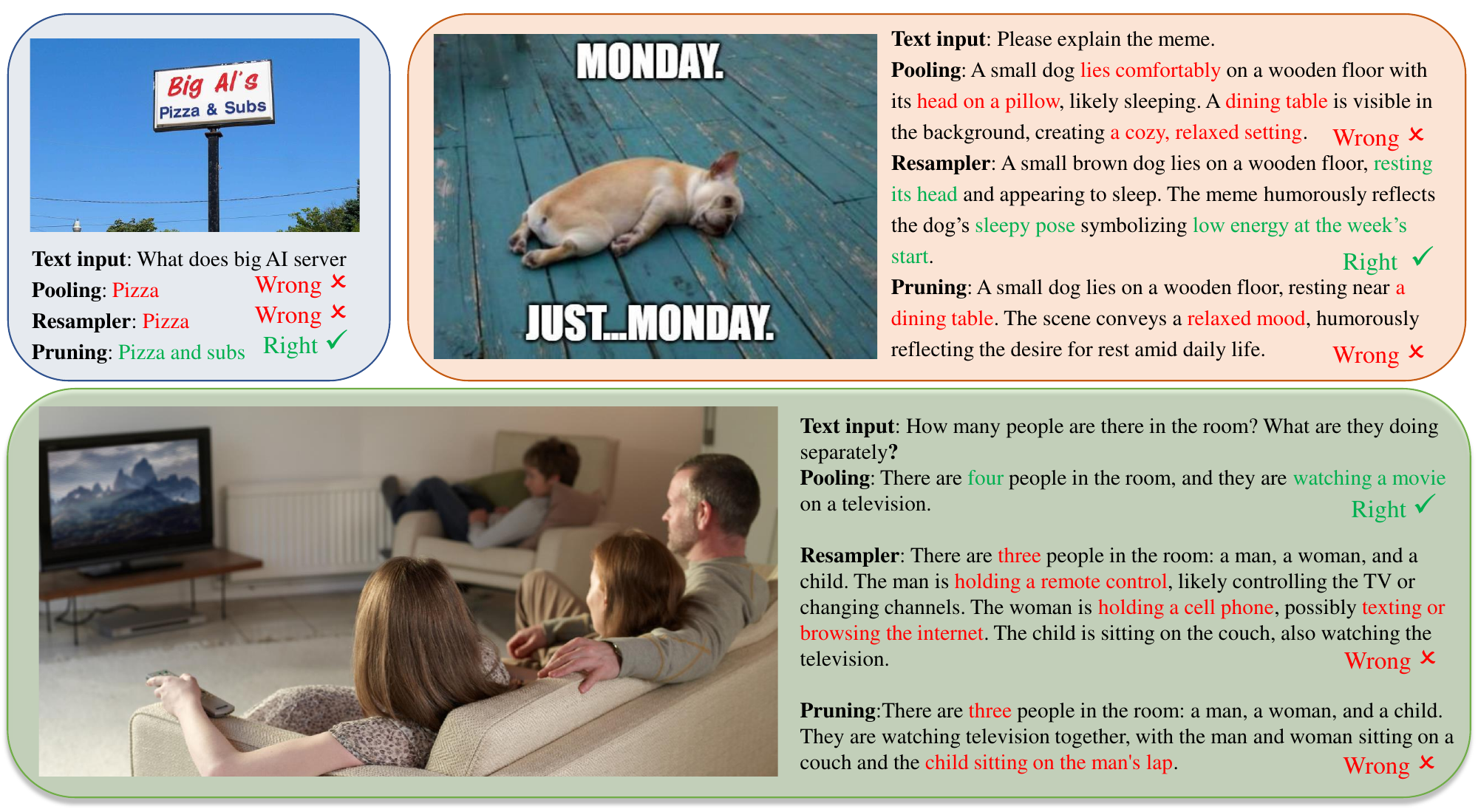}
  \caption{ Qualitative comparisons of three representative visual token compression methods under different scenarios. 
}
  \label{fig-2}
\end{figure*}
% 早期的方法采用简单的多层感知器（MLP）作为投影器，保持token数量不变，只在通道维度进行计算将视觉标记映射到文本嵌入空间（cite{llava15, minigpt2}）。然而，大量的视觉标记限制了这种方法的进一步发展。
% 最近的方法精心设计了投影器，不仅将视觉标记与文本特征对齐，还对视觉标记进行压缩。如图(a)、(b)和(c)所示，这些方法可分为三类：基于池化的方法、基于重采样器的方法和基于剪枝的方法。
% 基于pooling的方法（图\cite{7TP, 8LDP, 6honeybee}）应用卷积或最大汇集将空间上相邻的标记合并为紧凑的表示。基于重采样器的方法（QwenVL, Blip2）（图 c）引入了可学习的查询标记，并采用交叉注意机制从原始视觉标记中选择性地提取关键信息，从而有效地过滤和压缩高维特征。基于剪枝的方法 (图）评估标记的重要性并去除多余或不相关的标记，只保留信息量最大的标记用于下游配准。

Recent approaches have carefully designed sophisticated projectors that not only align visual tokens with textual features but also compress the visual tokens to reduce redundancy and computational cost. These methods can be categorized into three main strategies: pooling-based, resampler-based, and pruning-based approaches. 
Pooling-based methods \cite{7TP, 8LDP, 6honeybee} aggregate neighboring spatial tokens using operations such as convolution or max-pooling to produce coarse-grained, compact representations. Resampler-based methods \cite{QwenVL, Blip2} introduce learnable query tokens and employ cross-attention mechanisms to selectively extract key information from the original visual tokens, effectively filtering and compressing high-dimensional features. Pruning-based methods \cite{4FasterVLM, 9VisionZip} evaluate token importance and remove redundant or irrelevant tokens, retaining only the most informative ones for downstream alignment.

% 这些方法依赖于特定的压缩规则，这可能会限制不同任务和数据集之间的适应性。图\ref{fig-2}展示了三个典型案例，突出了每种方法的适用范围，说明了它们的优势和局限性。具体来说，基于汇集的方法保留了全局信息，但在压缩过程中对所有标记一视同仁，这可能会导致关键的局部细节丢失。基于重采样器的方法具有更大的灵活性，可以捕捉高级语义表征，但其复杂性会增加幻觉和丢失低级视觉线索的风险。基于剪枝的方法基于预定义的标准，专注于关键字元但可能导致重要信息被丢弃，并且导致视觉token的空间位置信息丢失。
However, these methods typically rely on specific and fixed compression rules, which limit their adaptability across different tasks and datasets. \cref{fig-2} presents three typical cases that illustrate the applicable scenarios, strengths, and limitations of each approach. Specifically, pooling-based methods are effective at preserving global context but treat all tokens uniformly, potentially discarding critical local details. Resampler-based methods offer greater flexibility in capturing high-level semantic representations through learnable queries and cross-attention mechanisms. Nonetheless, their increased complexity tends to introduce hallucinations and neglect low-level visual cues. 
Pruning-based methods focus on retaining key tokens based on predefined criteria, but may discard subtle yet important information and frequently disrupt spatial token structure.
\begin{figure}[t!]
  \centering
  \includegraphics[width=\linewidth]{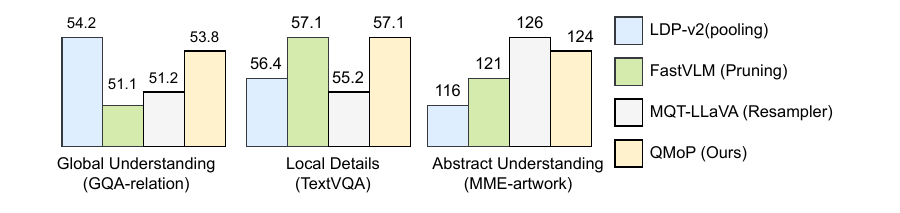}
  \caption{
  Comparison of compression strategies across tasks. Pooling excels on global understanding, pruning leads on local details, and resampling performs best on style and emotion, highlighting task-dependent advantages.
}
  \label{fig-1}
\end{figure}

% 除了定性结果，在现有的benchmark上的定性结果也能在一定程度上揭示某些压缩范式的局限性。例如，基于pruning的方法（例如FasterVLM、FastV）在需要细粒度文本识别的TextVQA上表现良好。此外，resampler在MME的“artwork”子集中表现出色（池化/修剪得分为126比118/117），这表明它在抽象场景理解方面具有优势。
Beyond qualitative observations, performance variations on existing benchmarks partially reflect the limitations of different compression strategies. As shown in \cref{fig-1}, pooling-based methods excel at capturing holistic scene semantics, achieving better results on relational reasoning tasks such as the relation subset of GQA. Pruning-based methods perform well on datasets like TextVQA, which require fine-grained textual recognition. Meanwhile, resampler-based methods achieve higher accuracy on abstract understanding tasks, scoring 126 on the artwork subset of MME. 
% 虽然现有的基准测试在一定程度上反映了不同压缩范式的局限性，但它们的信号往往是间接和零碎的。没有一个单一的基准可以对这三种策略进行全面的诊断，这严重阻碍了对视觉令牌压缩的深入分析和系统理解。
While existing benchmarks to some extent reflect the limitations of different compression paradigms, their signals are often indirect and fragmented. To the best of our knowledge, there is currently no benchmark that offers a comprehensive and unified diagnostic evaluation of pooling, resampling, and pruning strategies, which limits systematic analysis and impedes deeper understanding of visual token compression.

In this paper, we propose \textbf{Query-Guided Mixture-of-Projector} (\textbf{QMoP}), a flexible visual projection framework comprising three specialized branches: pooling‑based, resampler‑based, and pruning‑based. 
To adaptively combine these branches, we introduce a \textbf{Query-Guided Router} (\textbf{QGR}) that generates a weight for each projector based on both image and text inputs. 
Inspired by the Mixture‑of‑Experts paradigm, we then select and fuse only the two branches with the highest weights, discarding the least relevant one. This selective fusion not only emphasizes the most informative and complementary signals but also suppresses noise from less relevant projectors, enhancing the robustness of the final representation.
In addition, we propose \textbf{VTCBench}, a dedicated diagnostic benchmark designed to systematically evaluate the information degradation caused by Visual Token Compression (VTC). We conduct extensive analysis on VTCBench, validating our central hypothesis: different compression strategies lead to distinct forms of information loss and exhibit varying effectiveness across different task scenarios.
Our contributions are summarized as:
\begin{itemize}
% 我们
\item We conduct an empirical analysis of existing visual token compression strategies, identifying their inherent limitations and offering new insights for visual compression.
\item We introduce QMoP, a Query-Guided Mixture-of-Projector framework that dynamically integrates outputs from three complementary compression branches through a learned, context-aware routing module.
\item We propose VTCBench, a dedicated benchmark for visual token compression, which enables quantitative comparison across different compression paradigms. 
\end{itemize}

%% file: sec/2_related.tex
\section{Related Work}
\label{sec:formatting}
% 视觉令牌具有较高的空间冗余性和较低的信息密度,通常比文本令牌有着更多的数量。最近的一些方法也意识到了视觉令牌的冗余，并提出各种减少视觉令牌的方法。这些方法可以分为两类：在LLM解码过程中减少视觉token和在LLM之前减少视觉token。
Vision tokens exhibit high spatial redundancy and lower information density, typically outnumbering text tokens. Recent methods have recognized the redundancy in vision tokens and proposed various approaches to reduce their quantity. These methods can be divided into two categories: reducing vision tokens during the LLM decoding process and reducing vision tokens before they are fed into the LLM.

\subsection{Visual Token Reduction in LLM}
% 第一类方法在llm解码过程中减少vision token。FastV、VTK \cite{1FastV，3VTK}根据LLM某一层的文本-视觉注意力图修剪视觉标记。
% FitPrune \cite{5FitPrune}基于一小批数据来制定最佳的修剪策略，从而避免昂贵的手动试验。SparseVLM设计先验事先消除文本中的干扰，采用更准确的文本注意力剪枝视觉令牌
% 然而，由于 LLM 需要因果推理，因此在 LLM 中利用文本视觉注意力会导致注意力转移——位于图像右下角区域的视觉标记往往具有更高的注意力分数\cite{4FasterVLM}。
% 此外，LLM 中基于注意力的修剪需要额外的复杂操作来处理标记修改，这与高效的 Flash Attention 运算符相冲突，并且可以抵消通过减少标记序列长度所获得的速度增益 \cite{2502}。
Reducing vision tokens during the LLM decoding process has emerged as an effective avenue for improving computational efficiency.
FastV and VTK \cite{1FastV,3VTK} prune vision tokens based on attention scores at a specific LLM layer, retaining only the most informative tokens for downstream reasoning. To reduce manual tuning cost, FitPrune \cite{5FitPrune} leverages a small calibration set to derive optimal pruning ratios automatically. SparseVLM \cite{2SparseVLM} further incorporates prior knowledge to filter out distracting textual signals before applying attention-based token selection, thereby enhancing pruning accuracy. Beyond single-stage pruning, MustDrop and PDrop \cite{MustDrop,PDrop} adopt progressive token dropping across multiple inference stages. AutoPrune \cite{AutoPrune} dynamically adjusts pruning rates across layers based on task difficulty, while EPIC \cite{EPIC} progressively increases compression through distillation to maintain model performance at higher compression levels.
% 更进一步的， MustDrop在LLM内部分阶段逐步裁剪token，有着

\subsection{Visual Token Reduction in Visual Projector}
Recently, a growing body of work has explored compressing visual tokens within the projector while aligning visual and textual modalities. These approaches can be broadly categorized into resampler-based, pooling-based, and pruning-based paradigms.
\textbf{Pooling-based} projector compression reduces token count by spatially aggregating local visual patches. InternVL-1.5 \cite{InternVL15} employs pixel unshuffle to downsample tokens by expanding channel capacity. C-Abstractor \cite{6honeybee} and LDP \cite{8LDP} conduct learnable pooling via convolution kernels, while TokenPacker \cite{7TP} introduces localized attention to merge spatially adjacent features more adaptively.
\textbf{Resampler-based} designs extract salient information through cross-attention. Qwen-VL \cite{QwenVL} and BLIP-2 \cite{Blip2} employ learned queries to selectively attend to informative visual tokens, forming a content-aware bottleneck. LLaVA-Mini \cite{LLaVA-Mini} adopts a lightweight pre-fusion module that injects distilled visual signals directly into language tokens, offering a minimal-token fusion pipeline.
\textbf{Pruning-based} strategies instead remove redundant visual tokens after initial feature projection. FasterVLM \cite{4FasterVLM}, LLaVA-Prumerge \cite{LLaVAprumerge}, and VisionZip \cite{9VisionZip} rank tokens by CLS-based attention scores and preserve only the top-scoring subset under a fixed pruning ratio. SCOPE \cite{SCOPE} further alleviates semantic loss by incorporating set-coverage constraints derived from token–label relations, encouraging preservation of essential visual regions.
\subsection{Multimodal Benchmarks}
Existing multimodal benchmarks can broadly be categorized into two types. The first type focuses on high-level capabilities, such as reasoning, knowledge grounding, and complex task completion \cite{mmmu, mmmupro, mathvista, demon}. The second type, including MME \cite{mme}, MMStar \cite{mmstar}, and MMBench \cite{mmbench}, targets more fundamental perceptual and cognitive abilities through broad-spectrum evaluations.

While these benchmarks have greatly advanced the evaluation of MLLMs, they are not explicitly designed to assess the impact of visual token compression. In particular, their task formulations and categorizations do not isolate the visual fidelity loss caused by token reduction. Although some sub-tasks (e.g., fine-grained recognition, spatial reasoning) may partially reflect compression effects, the signals are indirect, entangled, and incomplete, making it difficult to diagnose and compare the behavior of different compression strategies in a principled manner.

%% file: sec/3_method.tex
\section{Method}
\begin{figure*}[!t]
  \centering
  \includegraphics[width=\linewidth]{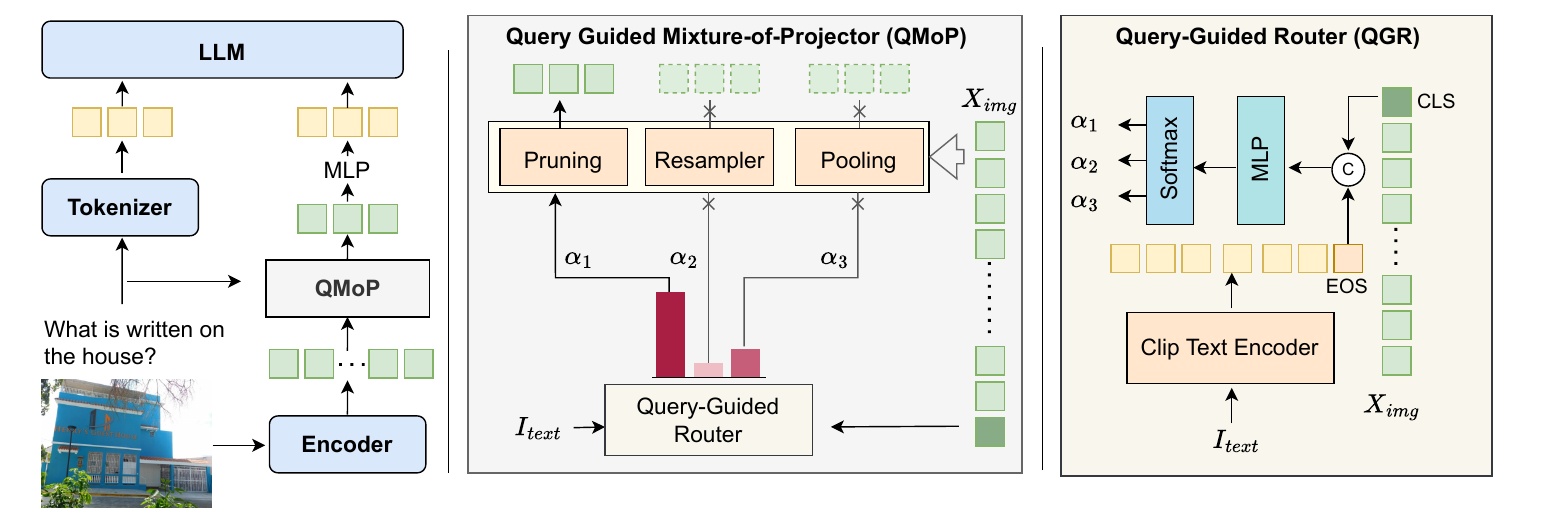}
  \caption{
    The architecture of the proposed QMoP.
    }
  \label{fig-3}
\end{figure*}
\subsection{Overview}
% 整体框架如图3所示，
% 输入的图像$I_{img}\in \mathbb{R}^{H\times W \times 3}$被分成了一个个patch后输入到视觉编码器生成视觉embeddings $X_img \in \mathbb{R}^{N \times C}$。接着projector根据X_img和输入的文本$I_{text}$自适应选择压缩方式,将X_img压缩成M个token对齐到文本空间以输出$T_img \in \mathbb{R}^{M \times C}$,其中M<N。视觉token生成的过程被制定为：
The overall framework is illustrated in \cref{fig-3}. Given an input image \(I_{img} \in \mathbb{R}^{H \times W \times 3}\), the image is first divided into a series of patches, which are then fed into the visual encoder to generate visual embeddings \(X_{img} \in \mathbb{R}^{N \times C}\), where \(N\) represents the number of patches and \(C\) is the embedding dimension. Next, the visual projector takes these visual embeddings along with the input text \(I_{text}\) and adaptively selects a compression strategy to reduce the number of visual tokens. In particular, \(X_{img}\) is compressed to produce a smaller set of tokens \(T_{img} \in \mathbb{R}^{M \times C}\), where \(M < N\), and these visual tokens are aligned with the text embedding space.
% 对于文本支路，$I_{text}$经过分词和编码后得到文本特征T_textemb，并与 T_img串联后输入到LLM，以自回归的形式输出相应。多模态生成内容的过程如下：
For the text branch, $I_{text}$ is tokenized and encoded to obtain text features $T_{text}$. The compressed visual tokens $T_{img}$ are then concatenated with $T_{text}$ and fed into the LLM, which generates the corresponding response in an autoregressive manner.

\subsection{Query Guided Mixture-of-Projector}
% 为了压缩视觉标记，我们提出了一种新颖的查询引导混合投影器（QMoP），它利用三种专门的压缩算子--池化、剪枝和重采样器--来捕捉视觉信息的不同方面。此外，QMoP 还集成了一种门控机制，可动态激活这些压缩分支的一部分并融合其特征，从而使该模型能够自适应地为每对图像-文本选择有效的压缩策略

To compress the visual tokens, we propose a novel Query Guided Mixture-of-Projector (QMoP) that harnesses three specialized compression operators-pooling, resampler, and pruning-to capture diverse aspects of visual information. In addition, QMoP integrates a gating mechanism that dynamically activates a part of these compression branches, enabling the model to select an effective compression strategy for each image-text pair adaptively.

% 为了有效利用池化、剪枝和重采样器分支的优势，我们引入了查询引导路由器（Query Guided Router），这是一种动态机制，可根据特定的输入图像和文本来决定激活哪种压缩策略。
To effectively harness the strengths of the pooling, pruning, and resampler branches, we introduce the Query Guided Router (QGR) as shown in \cref{fig-3}. QGR dynamically generates weighting coefficients that determine which compression strategies to activate based on the input image and text. 

For the visual branch, we extract the class token $v_{CLS} \in \mathbb{R}^{C_1}$ from the penultimate layer of the vision encoder. For the text branch, the input text \( I_{text} \) is fed into a pre-trained CLIP text encoder, from which we extract the EOS token $t_{EOS} \in \mathbb{R}^{C_2 }$.
These two tokens are then concatenated along the channel dimension to form a joint feature vector:
\begin{equation} 
f = \mathrm{Concat}(v_{CLS},\, t_{EOS}) \in \mathbb{R}^{(C_1+C_2)}.
\end{equation}

This combined feature \( f \) is passed through a two-layer multilayer perceptron (MLP) to generate the raw gating weights. This process can be formulated as:
\begin{equation} 
\{\alpha_1,\alpha_2,\alpha_3\} = \mathrm{softmax}(W_2 \cdot \sigma(W_1 f + b_1) + b_2),
\end{equation} 
where \(W_1 \in \mathbb{R}^{d \times (C_1+C_2)}\) and \(b_1 \in \mathbb{R}^{d}\) are the weights and bias of the first layer, \(W_2 \in \mathbb{R}^{3 \times d}\) and \(b_2 \in \mathbb{R}^{3}\) are those of the second layer, \(\sigma(\cdot)\) denotes a nonlinear activation function.
Finally, the output vector \(w \in \mathbb{R}^{3}\) is normalized using the softmax function to obtain the gating weights, which are subsequently used to control the contribution of each compression operator.

% % 大多数场景只需要灵活结合一到两个支路的信息已经足够了。融合三个支路不仅会增加计算量而且会引入冗余信息。相较于直接基于GQR的权重将token加权融合，我们提出了采用类似于MoE的方式舍弃最不重要的压缩分支，并将另外两个分支的结果：
% In most scenarios, flexibly combining information from only one or two branches is sufficient. Fusing all three branches not only increases computational overhead but also introduces redundant information. Instead of directly weighting and fusing tokens based on QGR scores, we propose an MoE-like strategy that discards the least important compression branch and fuses the remaining two branches' outputs based on their respective weights:
% \begin{equation}
% \begin{aligned}
% X_{\text{compression}} &= \alpha_i X_i + \alpha_j X_j, \\
% \alpha_i &= \max(\alpha_1, \alpha_2, \alpha_3), \\
% \alpha_j &= \max\left( \{\alpha_1, \alpha_2, \alpha_3\} \setminus \{\alpha_i\} \right),
% \end{aligned}
% \label{eq:compression_full}
% \end{equation}
% where \( X_i \) and \( X_j \) represent the outputs of the two selected branches, and \( \alpha_i \), \( \alpha_j \) are their corresponding gating weights.

\subsection{Compression Operator}
% 剪枝、resampler和池化算子分别基于不同的专家先验，将token数量为N的视觉特征压缩为M个token。在本文中他们的具体实现如下：
To reduce the computational burden while preserving task-relevant semantics, we design three token compression operators (pruning, resampling, and pooling), each driven by distinct expert priors. These operators transform the visual features with \(N\) tokens into a compact representation of \(M\) tokens. The detailed implementations of each operator are described as follows.

% 剪枝分支根据预定义规则为视觉标记分配分数，以剔除不重要的视觉标记。我们引入了两条规则：1.重要性，采用visual encoder最后一层的class token对其他token的Attention分数衡量图像中富含语义信息最丰富的区域；2.相关性，计算视觉token和文本的EOS之间的距离
\textbf{The pruning branch} assigns scores to visual tokens according to predefined rules to discard redundant ones. Specifically, we introduce two scoring criteria: (1) \textbf{Importance}, which measures how much attention the class token in the last layer of the visual encoder allocates to each visual token, thereby identifying regions that contribute most to semantic understanding; and (2) \textbf{Relevance}, which evaluates how closely each visual token is associated with the textual content by computing its distance to the EOS token in the text representation space. This process can be formulated as:  

% 我们具体来说，我们采用FasterVLM所定义的规则，取出视觉编码器最后一层的CLS attention(\(a_{CLS} \in \mathbb{R}^{N}\))。根据剪枝率计算动态阈值\tao并根据\tao减去不重要的分支。这一过程可表述如下
\begin{equation} 
s^{(i)} = \lambda s^{(i)}_{Importance} + (1-\lambda) s^{(i)}_{Relevance},
\end{equation} 
\begin{equation} 
X_{prune} = \{ X_{img}^{(i)} \mid s^{(i)} \geq \eta, \forall i \in [1, N] \},
\end{equation} 
where \(X_{img}^{(i)}\) represents the \(i\)-th visual token, $X_{prune}\in \mathbb{R}^{M \times C}$ is the refined token set that retains only the most informative tokens, $s^{(i)}$ is the importance score assigned to the \(i\)-th visual token according to predefined rules, \(\lambda\) is a hyperparameter controlling the balance between the two scoring criteria, which is set to 0.5 in our experiments. \(\eta\) is a dynamic threshold based on the pruning ratio. Tokens with scores below \(\eta\) are removed.

% 基于 Resampler 的压缩方法利用可学习的查询标记和交叉关注机制，从高维标记表征中选择性地提取和聚合关键视觉信息。
\textbf{The resampler branch} leverages learnable query tokens with cross-attention mechanisms to selectively extract and aggregate critical visual information from high-dimensional token representations. This process is defined as:  
\begin{equation} 
X_{resampler} = \text{Softmax} \left( \frac{QK^T}{\sqrt{d}} \right) V,
\end{equation} 
where $X_{resampler}\in \mathbb{R}^{M \times C}$ represents the resampled visual tokens, $Q\in \mathbb{R}^{M \times C}$ denotes the learnable query tokens, and $K, V\in \mathbb{R}^{N \times C}$ are the key and value representations derived from $X_{img}$.

% 池化分支将空间上相邻的标记汇总到一个紧凑的表示中，从而有效减少空间冗余，同时保留捕捉整体场景结构和物体间相对关系的基本视觉特征。为了在压缩过程中尽可能保持语义一致性与区域完整性，我们采用局部注意力机制进行池化。
\textbf{The pooling branch} aggregates spatially adjacent tokens into a compact representation, effectively reducing spatial redundancy while preserving essential visual cues that capture global scene structures and inter-object relations. To maintain semantic consistency and regional integrity during compression, we adopt a local attention mechanism for token pooling. 
% 首先将视觉特征\(X_{img} \in \mathbb{R}^{N \times C}\)reshape为二维的(X_{img2d} \in \mathbb{R}^{H \times W \times C}\),接着引入可学习的2d query (Q_{2d} \in \mathbb{R}^{h \times w \times C}\),其中N=H*W, M=h*w且H=sh,W=sw,s为下采样步长。Q_{2d}[i,j]仅与输入特征图中对应的局部窗口X_{img2d}[si:s(i+1),sj:s(j+1)]进行注意力交互，从而在空间上建立局部对齐关系。形式化的，局部注意力过程可写为：X_pool[i,j]=\text{Softmax} \left( \frac{Q_2d[i,j]K_{i,j}^T}{\sqrt{d}} \right) V_{i,j},其中K_{i,j}，V_{i,j}=X_{img2d}[si:s(i+1),sj:s(j+1)]
Specifically, the visual features \(X_{img} \in \mathbb{R}^{N \times C}\) are reshaped into a 2D feature map \(X_{img2d} \in \mathbb{R}^{H \times W \times C}\), where \(N = H \times W\). We then introduce a learnable 2D query map \(Q_{2d} \in \mathbb{R}^{h \times w \times C}\), where \(M = h \times w\) and \(H=s\cdot h, W=s\cdot w\), with \(s\) denoting the spatial downsampling stride. Each query \(Q_{2d}[i,j]\) interacts only with the corresponding local window \(X_{img2d}[si:s(i+1),sj:s(j+1)]\), thereby establishing a spatially aligned local aggregation. The local attention process is formulated as:
\begin{equation} 
X_{pool}[i,j]=\text{Softmax} \left( \frac{Q_{2d}[i,j]K_{i,j}^T}{\sqrt{d}} \right) V_{i,j},
\end{equation} 
where
\begin{equation} 
K_{i,j},V_{i,j} = \phi(X_{img2d}[si:s(i+1),sj:s(j+1)]),
\end{equation} 
and \(\phi(\cdot)\) denotes a linear projection.

\subsection{Training and Inference Strategy }
% QMoP遵循主流多式联运框架，采用两阶段培训方案。在第一阶段，仅训练投影仪将视觉特征映射到文本嵌入空间。本阶段不使用QG-Router；相反，三个专家分支的输出被连接并通过 MLP，允许所有分支共同学习到文本空间的投影。经过预训练后，每个分支获得初始的跨模态对齐能力
Following mainstream multimodal frameworks, QMoP adopts a two-stage training scheme. In the first stage, only the projector is trained to map visual features into the textual embedding space. In this stage, the QGR is not used; instead, the outputs from the three expert branches are concatenated and passed through an MLP, allowing all branches to learn the projection to the text space jointly. After pretraining, each branch acquires an initial cross-modal alignment capability.

% 在第二阶段，我们对投影仪和LLM进行联合微调，在此期间QG-Router被激活。在微调的早期阶段，我们为路由器设置了一个相对较大的温度系数\(\tau\)，以鼓励更平滑的路由分布。随着训练的进行，\(\tau\) 逐渐减少，以加强专家选择。为了增强多样性并防止路由器陷入次优局部最小值，在训练期间引入 Gumbel 噪声，并随着时间的推移逐渐退火。
In the second stage, we perform joint finetuning of the projector and the LLM, during which the QG-Router is activated. At the early phase of finetuning, we set a relatively large temperature coefficient \(\tau\) for the router to encourage smoother routing distributions. As training proceeds, \(\tau\) is gradually decreased to sharpen the expert selection. To enhance diversity and prevent the router from collapsing into sub-optimal local minima, Gumbel noise is introduced during training and gradually annealed over time.

% 为了稳定的梯度传播，我们在训练期间使用加权求和来聚合所有分支的输出。推理时，只有路由权重超过阈值的分支才会被激活；然后将它们的权重重新归一化，并通过激活分支的加权融合获得最终的视觉表示。
For stable gradient propagation, we aggregate the outputs of all branches using weighted summation during training. During inference, only branches whose routing weights exceed a threshold are activated; their weights are then re-normalized, and the final visual representation is obtained by a weighted fusion of the activated branches.
\begin{figure}[t!]
  \centering
  \includegraphics[width=0.85\linewidth]{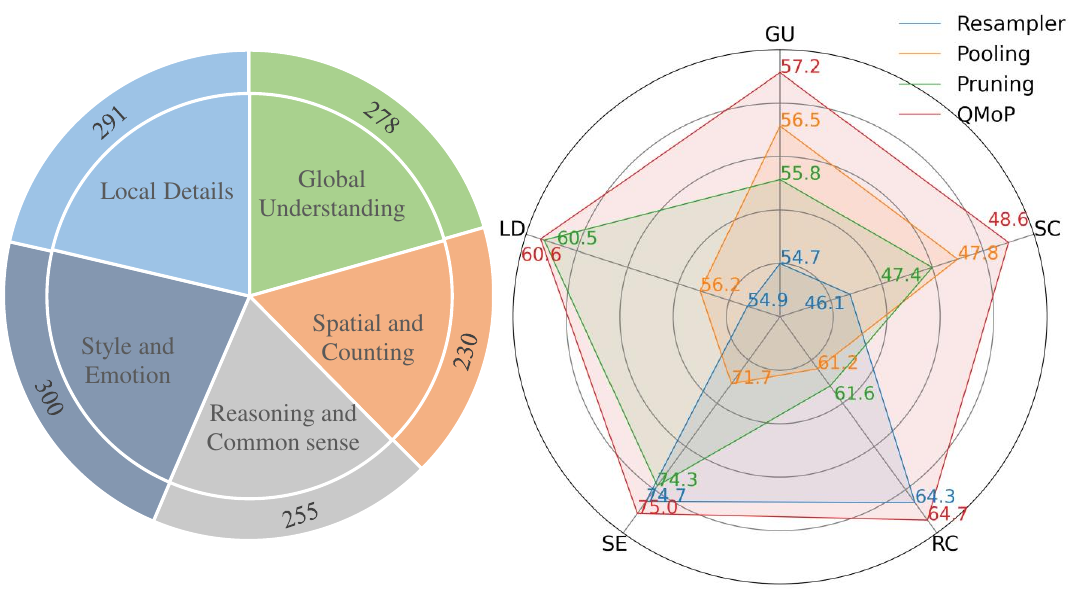}
  \caption{
 Overview of VTCBench and the quantitative analysis of compression strategies.
}
  \label{fig-VTC}
\end{figure}
\begin{table}[t!]
  \centering
  \caption{Performance comparison with pre-LLM compression methods on LLaVA-1.5-7B. The best and second-best results are highlighted in \textbf{bold} and \underline{underlined}.}
  \setlength{\tabcolsep}{1pt} 
  % \scriptsize 
  \fontsize{7pt}{9pt}\selectfont
  \begin{tabular}{l | c c c c c c c c c c | c}
    \hline\hline
    Methods  & VQA\textsuperscript{T} & POPE & VQA\textsuperscript{v2} & MM-Vet & 
    GQA & MME & MMB & Seed & MMStar & MMMU & Avg.\\
    \hline
    \rowcolor{gray!10}
    \multicolumn{12}{c}{ LLaVA-1.5-7B, Vision token = 144}\\
    \hline
    Vanilla \cite{llava15}  & 58.2 & 85.9 	& 78.5 & 31.6 &	62.0 &	1511 &	64.3 &	66.1 &	31.9 &	34.6 &	100.00\%\\
    Pixel-Shuffle \cite{InternVL15} & 54.2 &	87.0 & 75.6 &	29.1 &	60.2 &	1422 &	64.5 &	61.6 &	31.7 &	32.7 &	96.14\% \\
    FasterVLM \cite{4FasterVLM} & \underline{57.1} &	83.5 &	76.2 &	32.2 &	58.0 &	1440 &	64.3 &	62.3 &	31.3 &	\underline{34.8} &	97.64\%\\
    C-Abstractor \cite{6honeybee} & 54.1 	& 85.9 &	75.7 &	28.2 &	60.4 &	1392 &	63.3 & 	62.3 &	31.1 &	33.4 &	95.48\%\\
    MQT-LLaVA \cite{14MQT} & 52.6 &	86.2 &	76.8 &	29.8 &	61.4 &	1445 &	64.4 & 62.0 &	31.9 &	34.8 &	97.22\% \\
    TokenPacker \cite{7TP}& 57.0 & \underline{87.0} &	\underline{77.9} &	\underline{33.0} &	\underline{61.9} &	\underline{1454} &	\textbf{65.0} &	\underline{65.1} &	\underline{32.6} &	34.3 &	\underline{100.05\%} \\
    LDP-v2 \cite{8LDP}    & 56.4 &	86.3 &	77.3 &	31.7 &	61.5 &	1435 &	65.1 & 63.9 &	31.9 &	33.9 &	98.62\%\\
    QMoP (ours)& \textbf{57.1} &	\textbf{87.7} &	\textbf{78.1} &	\textbf{33.3} &	\textbf{62.0} &	\textbf{1456} &	\underline{64.8} &	\textbf{66.0} &	\textbf{33.1} &	\textbf{35.1} &	\textbf{100.73\%}\\
    \hline
    \rowcolor{gray!10}
    \multicolumn{12}{c}{ LLaVA-1.5-7B, Vision token = 64}\\
    \hline
    Pixel-Shuffle \cite{InternVL15} & 52.2 &	85.5 &	73.2 &	25.8 &	58.7 &	1383 & 64.4 & 	59.3 &	30.1 &	34.6 &	93.46\% \\
    FasterVLM \cite{4FasterVLM} & \underline{56.0} &	80.4 &	72.5 &	28.7 &	55.0 &	1342 &	61.6& 	57.9 &	\underline{31.8} &	\underline{34.6} &	93.37\% \\
    C-Abstractor \cite{6honeybee} & 51.3 &	85.7 &	73.7 &	26.1 &	59.1 &	1415 &	61.8 &	60.0 &	31.3 &	34.1 &	93.69\% \\
    MQT-LLaVA \cite{14MQT} & 51.6 &	83.6 &	75.3 &	28.9 &	60.0 &	\underline{1454} &	63.5 &	60.8 &	30.9 &	34.4 &	95.41\% \\
    TokenPacker \cite{7TP}& 55.6 &	86.5 &	\textbf{77.2} &	29.1 & \textbf{61.1} &	1440 &	64.4 &	\underline{63.0} &	31.3 &	33.8 &	97.18\% \\
    LDP-v2 \cite{8LDP}    & 55.1 &	\underline{86.8} &	76.1 &	\underline{30.1} &	60.1 &	1414 &	\textbf{66.2} &	62.6 &	31.5 &	34.2 &	\underline{97.38\%} \\
    QMoP (ours) & \textbf{56.0} &	\textbf{87.2} &	\underline{76.4} &	\textbf{30.8} &	\underline{60.2} &	\textbf{1455} &	\underline{64.6} &	\textbf{63.6} &	\textbf{31.9} &	\textbf{34.6} &	\textbf{98.25\%}\\
    \hline
    \rowcolor{gray!10}
    \multicolumn{12}{c}{ LLaVA-1.5-13B, Vision token = 144}\\
    \hline
    Vanilla \cite{llava15}  & 63.8 &	87.9 &	80.0 &	35.4 &	63.3 &	1531 &	67.7 &	68.2 &	33.1 &	36.1 &	100.00\% \\

    Pixel-Shuffle \cite{InternVL15} & 57.0 &	86.9 &	76.9 &	33.1 &	61.2 &	1483 &	66.2 &	64.0 &	32.1 &	34.4 &	95.53\% \\
    FasterVLM \cite{4FasterVLM} & \underline{58.8} &	86.1 &	77.1 &	33.6 &	59.1 &	1489 &	66.4 &	63.8 &	32.3 &	34.9 &	95.79\%\\
    C-Abstractor \cite{6honeybee} & 53.9 &	86.3 &	74.8 &	28.2 &	60.4 &	1391 &	63.0 &	61.0 &	31.3 & 35.4 &	91.73\%\\
    TokenPacker \cite{7TP}& 58.8 & 87.4 & \underline{78.8} & \underline{34.5} & \underline{62.3} &	\underline{1525} & \underline{67.4} &	\underline{65.4} &	\textbf{33.9} &	35.0 &	\underline{98.06\%}\\
    LDP-v2 \cite{8LDP}    & 58.4 &	\underline{87.6} &	78.4 &	34.4 &	62.0 &	\textbf{1533} & 66.9 & 65.3 &	33.2 &	\underline{35.6} & 	97.80\%\\
    QMoP (ours)& \textbf{59.4} &	\textbf{88.2} &	\textbf{79.4} &	\textbf{34.7} &	\textbf{62.7} &	1519 & \textbf{67.4} &	\textbf{67.0} &	\underline{33.7} &	\textbf{36.1} &	\textbf{98.86\%}\\
    \hline\hline
  \end{tabular}
  \label{tab-1}
\end{table}
\subsection{Benchmark for Visual Token Compression}
% 为了更全面地分析丢弃视觉令牌的影响，我们开发了\textbf{VTCBench}，这是一个专门的基准测试，旨在评估视觉令牌压缩引起的功能退化。VTCBench 从五个关键维度来衡量标记压缩性能，每个维度都能捕捉到视觉信息保存的一个不同方面：
To enable a more comprehensive analysis of the impact caused by discarding visual tokens, we introduce \textbf{VTCBench}, a dedicated benchmark designed to evaluate functional degradation induced by visual token compression. VTCBench measures compression performance across five key dimensions:
1) \textbf{Global Understanding (GU)}: Evaluates the model’s ability to retain the overall scene context, ensuring that essential background and semantic information remain intact after compression.
2)\textbf{Spatial and Counting (SC)}: Assesses the model’s capacity to preserve spatial relationships and accurately count objects.
3) \textbf{Reasoning and Common sense (RC)}:  Tests whether the compressed representations still support higher-order inference, including logical reasoning and commonsense knowledge grounded in the visual content.
4) \textbf{Style and Emotion (SE)}: Examines the preservation of artistic and affective elements, such as color composition, texture, and emotional expressions conveyed through images.
5) \textbf{Local Details (LD)}: Focuses on the preservation of fine-grained visual features such as text, small objects, or intricate patterns, which are easily lost under aggressive token reduction.

% CPBench 的构建过程分为多个阶段。首先，我们选择了六个既涵盖自然图像又涵盖特定领域场景的现有数据集（\cite{mme, textvqa, gqa, seed, mmstar, sqa}）。接下来，我们将应用基于 LLM 的专有自动过滤管道（\cite{gpt4}）来移除不符合我们五个评估维度的样本。在使用 LLM 检测器进行初步粗过滤后，我们会进行人工审核，以纠正任何错误分类。最后，考虑到评估维度和任务难度的多样性，我们从过滤后的样本库中手动筛选出超过 1000 个均衡的高质量样本。这些样本的分布全面覆盖了所有五个维度。图~ref{fig-33}提供了 CPBench 构成的详细分类。
The construction of VTCBench follows a multi-stage process. First, we select six existing datasets that cover both natural images and domain-specific scenarios (\cite{mme, textvqa, gqa, seed, mmstar, sqa}). Next, we apply a proprietary LLM-based automatic filtering pipeline \cite{gpt4} to remove samples that do not align with our five evaluation dimensions. After an initial coarse filtering using an LLM detector, we perform a manual review to correct any misclassifications. Finally, taking into account the diversity of evaluation dimensions and task difficulty, we manually curate a balanced set of over 1,000 high-quality samples from the filtered pool. These samples are distributed to cover all five dimensions comprehensively. \cref{fig-VTC} provides a detailed breakdown of the composition of VTCBench. Please refer to the supplementary material for more details.

%% file: sec/4_experiment.tex
\section{Experiments}
\subsection{Implementation Details}
% 我们选择LLaVA-1.5-7B 作为baseline，验证QMoP的有效性。我们采用与LLaVA-1.5-7B相同的训练数据与超参数设置。此外
We chose LLaVA-1.5-7B \cite{llava15} as our primary baseline to evaluate the effectiveness of QMoP. All experiments are conducted by pretraining on the LAION-CC-SBU-558K dataset and performing instruction tuning on a 665K mixed dataset. We employ Vicuna-7B as the LLM and CLIP-ViT-Large-Patch14-336 as the visual encoder.  We fix the number of visual tokens to 144 and 64 to ensure fair comparisons with previous methods under similar computational cost.
To further assess the scalability and robustness of QMoP, we evaluate it on LLaVA-1.5-13B (larger model size). 
% 我们通过 10 个具有代表性的公开视觉理解基准和我们制作的VTCBench来评估我们的模型。有关基准测试的更多详细信息，请参阅附录
We evaluate our model on ten representative public vision-understanding benchmarks, as well as on our newly constructed VTCBench. For additional details about these evaluation benchmarks, please refer to the supplementary material for more details.

\begin{table*}[t!]
  \centering
  \caption{Performance comparison with intra-LLM visual token compression methods on LLaVA-1.5-7B.}
  \setlength{\tabcolsep}{1.5pt} 
  \fontsize{8pt}{10pt}\selectfont
  \begin{tabular}{l|c | c c c c c c c c}
    \hline\hline
     \multirow{2}{*}{Method} & Vision  & \multirow{2}{*}{VQA\textsuperscript{T}} & \multirow{2}{*}{POPE} & \multirow{2}{*}{VQA\textsuperscript{v2}}& \multirow{2}{*}{GQA}& \multirow{2}{*}{MMB} & MMB- & OCR & \multirow{2}{*}{Avg.}  \\
     & Token &  & & & & & CN& Bench\\
    \hline
    LLaVA-1.5-7B & 576 &	58.2 	&85.9 &	78.5 &	62.0 &	64.3 &	58.3 &	297.0 &	100.00\%\\
    FastV \cite{1FastV} & 192  & 52.5 & 64.8  & 67.1 & 52.7  & 61.2 & 57.0 &	291 & 89.58\% \\
    HiRED \cite{HiRED} & 192 &	47.4	& 82.8 & 74.9	& 58.7	& 62.8	& 54.7	& 190 &	89.06\% \\
    FitPrune \cite{5FitPrune} & 192 &	57.4 &	83.4 & - & \underline{60.4}	& 63.3	& 56.4	& -	& - \\
    LLaVA-PruMerge \cite{LLaVAprumerge}	& 192& 	54.3	&71.3	&70.6	&54.3	&59.6	&52.9	&253 &	88.92\% \\
    SparseVLM \cite{2SparseVLM} & 192 &	56.1 &	\underline{83.6} & 75.6	& 57.6 & 62.5 & 53.7 & 292 & 95.79\% \\
    PDrop \cite{PDrop} & 192 & 56.1 & 82.3 & 75.1 & 57.1 & 63.2 & 56.8 & 290 & 96.19\% \\
    MustDrop \cite{MustDrop} & 192 & 56.5 & 82.6 & 76 & 58.2 & 62.3	& 55.8	& 289 & 96.26\% \\
    DART \cite{DART} & 192  & \textbf{57.4} &	82.8 & \underline{76.7} & 60	& \underline{63.6}	& \underline{57.0} & \underline{296} & \underline{97.98\%} \\
    QMoP (Ours) & 144 & \underline{57.1} & \textbf{87.7} & \textbf{78.1} & \textbf{62.0} & \textbf{64.8} &	\textbf{57.0} & \textbf{304} & \textbf{100.08\%} \\
    \hline\hline
  \end{tabular}
  \label{tab-2}
\end{table*}
\subsection{Comparison on Mainstream Benchmarks}
% 我们将QMoP与其他visual token压缩方法在主流数据集上进行全面的比较。具体来说，比较的方法包括在LLM之前压缩的方法如：FasterVLM、 MQT-LLaVA、Pixel-Shuffle、C-Abstractor、TokenPacker和LDP-v2，以及在LLM内部压缩的方法包括FastV、LLaVolta、PDrop。
We conduct a comprehensive comparison between QMoP and existing visual token compression methods across several standard benchmarks. Specifically, we include both pre-LLM compression methods: FasterVLM \cite{4FasterVLM}, MQT-LLaVA \cite{14MQT}, Pixel-Shuffle \cite{InternVL15}, C-Abstractor \cite{6honeybee}, TokenPacker \cite{7TP}, and LDP-v2 \cite{8LDP}, as well as intra-LLM compression methods, include FastV \cite{1FastV}, HiRED \cite{HiRED}, FitPrune \cite{5FitPrune}, LLaVA-PruMerge \cite{LLaVAprumerge}, SparseVLM \cite{2SparseVLM}, PDrop \cite{PDrop}, MustDrop \cite{MustDrop} and DART \cite{DART}.

% 如表\ref{tab-1}所示，不同的压缩策略在各种基准测试中表现出明显的优势。在像TextVQA\cite{TextVQA}这样严重依赖局部细节进行文本识别的任务中，基于修剪的方法，如FastV\cite{1FastV}、FitPrune \cite{5FitPrune}和FasterVLM\cite{4FasterVLM}，通过选择性地保留细粒度的视觉线索来实现强大的性能。对于更全面的推理任务，如MME\cite{MME}，基于重采样器的方法，如MQT LLaVA\cite{14MQT}，表现更好，这可能是因为它们能够提取高级语义表示。相比之下，基于池的方法，如TokenPacker\cite{7TP}和LDP-v2\cite{8LDP}，在VQAv2\cite{VQAv2}、ScienceQA\cite{sqa}和GQA\cite{GQA}等通用QA基准上显示出具有竞争力的结果，受益于它们通过空间聚合来保存全局上下文的能力。
% 我们的方法QMoP根据输入图像和文本上下文动态选择压缩策略。这种灵活性使其能够在所有评估的基准测试中实现最高或接近最高的性能。
For pre-LLM compression methods, we compress the number of visual tokens to 144 and 64 to ensure fair comparisons.
As shown in \cref{tab-1}, different compression strategies excel on different tasks. Pruning-based methods (e.g., FasterVLM \cite{4FasterVLM}) perform well on TextVQA \cite{textvqa} by retaining fine-grained visual details. Resampler-based methods (e.g., MQT-LLaVA \cite{14MQT}) are better suited for reasoning-heavy tasks like MME \cite{mme}, due to their semantic abstraction capabilities. Pooling-based methods (e.g., TokenPacker \cite{7TP}, LDP-v2 \cite{8LDP}) achieve strong results on VQAv2 \cite{vqav2} and GQA \cite{gqa} by preserving global context through spatial aggregation. In contrast, QMoP dynamically selects the compression strategy based on the input image and text context. This flexibility enables it to achieve top performance across all evaluated benchmarks.

% 为了进行更全面的评估，我们在表2中进一步将我们的方法与LLM内压缩方法进行了比较。这些方法侧重于减少语言模型解码过程中的视觉标记，而不是在投影仪级别。如结果所示，即使在更高的压缩比（144比192个令牌）下，我们的方法在多个基准测试中也能实现始终如一的卓越性能。这表明，我们的投影仪不仅在积极压缩下保留了基本的视觉语义，而且比LLM本身中压缩令牌的方法保持了更强的跨模态对齐。
For a more comprehensive evaluation, we further compare our approach with intra-LLM compression methods in \cref{tab-2}. These methods focus on reducing visual tokens within the language model’s decoding process rather than at the projector level. As the results show, even under a higher compression ratio (144 vs. 192 tokens), our method achieves consistently superior performance across multiple benchmarks. 
% This demonstrates that our projector not only preserves essential visual semantics under aggressive compression but also maintains stronger cross-modal alignment than methods compressing tokens within the LLM itself.
\begin{table}[t!]
\setlength{\tabcolsep}{6pt} %调整列间距，默认是6
    \centering
    \caption{Performance comparison on VTCBench. The results are reported across five dimensions: GU (Global Understanding), SC (Spatial and Counting), RC (Reasoning and Common Sense), SE (Style and Emotion), and LD (Local Details). A, B, and C indicate whether the method is based on resampler, pooling, or pruning, respectively.}
  \fontsize{8}{10}\selectfont
    \begin{tabular}{c| c | c c c c c| c}
    \hline
    \hline
     Method   & Type &  GU & SC & RC & SE & LD & Avg.\\
     \hline
     LLaVA-1.5-7B \cite{llava15} & - & 56.1 & 47.4 &	65.9 &	72.0 &	60.9 &	61.0 \\
     \hline
     MQT-LLaVA \cite{14MQT} & A & 54.7 &	46.1 &	64.3 &	\underline{74.7} &	54.9 &	59.5  \\
     Pixel-Shuffle \cite{InternVL15} & B & 53.6 &	47.0 & 	61.2 &	70.0 &	55.3 &	57.9  \\
     C-Abstract \cite{6honeybee} & B &53.2 &	47.8 &	59.6 &	70.3 &	55.2 &	57.7  \\
     TokenPacker \cite{7TP} & B & 55.8 & 47.8 &	\underline{64.7} &	74.0 &	57.5 &	60.5 \\
     LDP-V2 \cite{8LDP} & B & \underline{56.8} &	\underline{48.3} &	63.9 &	74.3 &	56.4 & \underline{60.5}  \\
     FasterVLM \cite{4FasterVLM} & C & 54.7 &	43.5 &	59.6 &	70.7 &	\underline{58.3} &	58.0 \\
     QMoP & MoE & \textbf{57.2} &	\textbf{48.6} &	\textbf{64.7} &	\textbf{75.0} &	\textbf{60.6} &	\textbf{61.8} \\
     \hline
     \hline
    \end{tabular}
    \label{tab-VTC}
\end{table}

\subsection{Comparison on VTCBench}
% 为了全面评估标记压缩方法的有效性，我们将 QMoP 与现有的标记压缩方法在 CPBench 的所有五个维度上进行了比较。表ref{tab-2}报告了不同方法在CPBench上的准确度，得分越高表示视觉信息保留得越好。
To comprehensively evaluate the effectiveness of token compression methods, we compare QMoP with existing token compression methods across all five dimensions of VTCBench. Table \ref{tab-VTC} reports the accuracy of different methods, where a higher score indicates better retention of visual information.

As a \textbf{resampler-based} projector, \textbf{MQT-LLaVA} \cite{14MQT} exhibits notable advantages in the \textit{Reasoning and Common Sense} and \textit{Style and Emotion} dimensions, despite its weaker performance in other aspects.
As \textbf{pooling-based} methods, \textbf{TokenPacker} \cite{7TP} and \textbf{LDPv2} \cite{8LDP} demonstrate strong performance in \textit{Global Understanding} and \textit{Spatial and Counting}, benefiting from their ability to retain coarse-grained global context.  However, these methods apply uniform compression across all tokens without considering their importance, leading to significant performance drops in \textit{Local Details}. In contrast, the \textbf{pruning-based} method \textbf{FasterVLM} \cite{4FasterVLM} achieves competitive performance on \textit{Local Details}, ranking second only to our proposed QMoP.

QMoP dynamically selects appropriate compression strategies according to the input, effectively balancing the retention of global context and the preservation of fine-grained details. As a result, it achieves consistently strong performance across all five dimensions of VTCBench.

\begin{table}[t!]
    \setlength{\tabcolsep}{6pt} %调整列间距，默认是6
    \centering
    \caption{Ablation study on different projector combinations across VTCBench.  The results are reported across five dimensions: GU (Global Understanding), SC (Spatial and Counting), RC (Reasoning and Common Sense), SE (Style and Emotion), and LD (Local Details). A, B, and C denote the three compression methods: resampler, pooling, and pruning, respectively.}
     {
    \fontsize{8}{10}\selectfont 
    \begin{tabular}{c|c c c c c| c}
    \hline
    \hline
     Method    &  GU & SN & RC & SE & LD & Avg.\\
     \hline
     LLaVA-1.5-7B& 56.1 & 47.4 &	65.9 & 	72.0 &	60.9 &	61.0 \\
     \hline
     A & 54.7 &	46.1 &	\underline{64.3} &	\underline{74.7} &	54.9 &	59.5  \\
     B & 56.5 &	\underline{47.8} &	61.2 &	71.7 &	56.2 &	59.2  \\
     C & 55.8 &	47.4 &	61.6 &	74.3 &	\underline{60.5} &	60.3  \\
     A+B & 56.8 & 47.8 &	63.1 &	72.3 &	57.6 &	60.1 \\
     A+C & 57.2 & 47.8 &	62.7 &	73.0 &	59.2 &	60.4 \\
     B+C& \textbf{58.6} &	47.4 &	61.6 &	73.0 &	59.6 &	\underline{60.9}  \\
     A+B+C & 55.8 &	47.4 &	60.8 &	74.3 &	59.6 &	60.2 \\
     QMoP & \underline{57.2} &	\textbf{48.6} &	\textbf{64.7} &	\textbf{75.0} &	\textbf{60.6} &	\textbf{61.8}\\
     \hline
     \hline
    \end{tabular}
    }
    \label{tab-3}
\end{table}

\subsection{Ablation Study}
\subsubsection{Base Projector Combination}
% 为了研究不同压缩策略组合的有效性，我们进行了一项消融研究，评估各种投影仪组合如何影响 CPBench 五个评估维度的压缩性能。具体来说，我们分析了基于池化、基于剪枝和基于重采样器的三种基本投影仪的不同子集如何影响信息保留。我们在禁用查询引导路由器（QGR）的同时，有选择性地同时启用一个或两个投影器，以隔离每种压缩策略的影响。评估的组合包括：
To investigate the effectiveness of different compression strategy, we conduct an ablation study by combining different compression strategy.
Specifically, we selectively enable some projectors at a time while disabling the Query Guided Router (QGR) to isolate the impact of each compression strategy.
\cref{tab-3} presents the performance of different projector combinations across VTCBench. Our key observations are as follows:
\begin{itemize}
\item Resampler-based compression exhibits superior high-level representation learning and effectively extracts latent semantic features. 
\item Pooling-based compression demonstrates strong performance in global understanding, effectively preserving overall scene context. 
\item Pruning-based compression improves local detail. However, its rigid selection criteria lead to suboptimal performance in capturing high-level semantics and contextual relationships. 
\item Directly merging all three outputs introduces noise, often leading to worse performance. A selective mechanism is thus crucial.
\end{itemize}
\begin{table}[t!]
    \setlength{\tabcolsep}{6pt} %调整列间距，默认是6
    \centering
    \caption{Ablation on the number of visual tokens in QMoP.}
     {
    \fontsize{8}{10}\selectfont 
    \begin{tabular}{c| c c c c c c c}
    \hline
    \hline
    Tokens & VQA\textsuperscript{T} & VQA\textsuperscript{v2} & GQA & MMB & POPE & MM-Vet & Avg.(\%)\\
     \hline
     576 & 58.2 & 78.5 & 62.0 & 64.3 & 85.9 & 31.6 & 100.00\\
     \hline
     144 & 57.1	& 78.1 & 62.0 & 64.8 & 87.7 & 33.3 & 99.49\\
     64  & 56.0 & 76.4 &	60.2 &	64.6 &	87.2 & 30.8 & 96.91 \\
     36  & 54.2 & 74.1 & 58.4 & 62.3 &	86.8 & 28.7 & 93.67\\
     16 & 51.0 & 73.2 & 58.2 & 61.3 & 86.5 & 26.3 & 92.34 \\
     4 &	51.1 & 70.8 & 56.7 & 59.6 & 82.9 & 26.4 & 89.03\\
     \hline
     \hline
    \end{tabular}
    }
    
    \label{tab-num}
\end{table}

\subsubsection{The Number of Compressed Visual Tokens}
To investigate the impact of different compression ratios on model performance, we conduct an ablation study by varying the number of visual tokens retained after compression. Specifically, we evaluate QMoP with 4, 16, 36, 64, and 144 tokens on several representative multimodal benchmarks, including TextVQA \cite{textvqa}, VQAv2 \cite{vqav2}, GQA \cite{gqa}, MMBench \cite{mmbench}, POPE \cite{pope}, and MM-Vet \cite{mmvet}.
% 如表n所示，LLaVA的token数量为576，我们压缩到144，整体性能基本不变，在一些benchmark还有着超越baseline的表现。这说明……。而随着压缩率进一步上升，……整体性能逐渐下降。
As shown in \cref{tab-num}, the baseline LLaVA employs 576 visual tokens, whereas QMoP compresses them to only 144. Remarkably, the overall performance remains nearly unchanged, and on several benchmarks, QMoP even surpasses the baseline. This demonstrates that QMoP effectively removes redundant visual information while preserving the most informative semantic cues crucial for cross-modal reasoning. However, as the compression ratio further increases (e.g., 64 to 4 tokens), the overall performance gradually declines, suggesting that excessive compression inevitably results in the loss of fine-grained visual details necessary for complex understanding.
\begin{table}[b!]
    \setlength{\tabcolsep}{6pt} %调整列间距，默认是6
    \centering
    \caption{Complexity comparison between QMoP and baseline (LLaVA-1.5-7B).}
     {
    \fontsize{8}{10}\selectfont 
    \begin{tabular}{c|c c c c c}
    \hline
    \hline
    \multirow{2}{*}{Tokens} & \multirow{2}{*}{TFLOPs} & \multirow{2}{*}{KVcache} & Train & Inference  & \multirow{2}{*}{Performance} \\
    & & & time & time & \\
    \hline
    576 & 3.82 & 302.0M & 102.3h & 3h05m & 100\%\\
    \hline
    144 & 0.94 & 75.5M  & 75.6h  & 2h29m & 100.73\%\\
     64 & 0.42 & 33.6M  & 65.5h  & 2h20m & 97.97\%\\
     36 & 0.23 & 18.9M  & 61.3h  & 2h20m & 94.97\%\\
     16 & 0.10 & 8.4M   & 58.0h  & 2h18m & 92.78\%\\
     4 & 0.03 & 2.1M   & 56.0h    & 2h17m & 91.70\%\\
    \hline
    \hline
    \end{tabular}
    }
    
    \label{tab-efficient}
\end{table}
\subsubsection{Complexity Analysis}
To comprehensively evaluate the efficiency–performance trade-off of QMoP under different compression ratios, we compare it with the uncompressed baseline LLaVA-1.5-7B across five key dimensions:
\begin{itemize}
    \item \textbf{Computational cost}, measured by the total FLOPs required during inference;
    \item \textbf{Memory efficiency}, quantified by the KV cache size used for storing visual features;
    \item \textbf{Training time}, representing the end-to-end time required for model optimization under the same config;
    \item \textbf{Inference time}, measured as the total response latency on the VQAv2 \cite{vqav2} test set;
    \item \textbf{Average performance}, defined as the mean normalized accuracy across ten representative multimodal benchmarks, including TextVQA \cite{textvqa}, POPE \cite{pope}, VQAv2 \cite{vqav2}, MM-Vet \cite{mmvet}, GQA \cite{gqa}, MME \cite{mme}, MMBench \cite{mmbench}, Seed-IMG \cite{seed}, MMStar \cite{mmstar}, and MMMU \cite{mmmu}.
\end{itemize}

As shown in \cref{tab-efficient}, when the number of visual tokens is reduced from 576 (baseline) to 144, QMoP achieves substantial efficiency gains, consuming only 25\% of the baseline FLOPs and KV cache memory, while maintaining 100.73\% of the baseline’s average performance. This suggests that moderate compression effectively removes redundant visual information while retaining the essential semantic cues needed for robust cross-modal reasoning.
However, as the compression ratio further increases to 36 tokens, the average performance drops sharply, even though the reductions in training and inference time become marginal. This indicates a compression threshold, beyond which excessive token removal leads to significant loss of visual semantics and degraded overall reasoning capability.

\begin{table}[b!]
    \setlength{\tabcolsep}{6pt} %调整列间距，默认是6
    \centering
    \caption{Efficiency breakdown of the projector overhead.}
    {\fontsize{8}{10}\selectfont
        \begin{tabular}{l | c c c c}
            \hline
            Method & TokenPacker & LDP-V2 & QGR & QMoP \\
            \hline
            FLOPs & 10.73G & 12.1G & 3.57G & 7.29G+3.57G \\
            \hline
        \end{tabular}
    }
    \label{tab-R2}
\end{table}
\subsection{Efficiency breakdown of the projector overhead.}
Table~\ref{tab-R2} reports the FLOPs overhead of the projector modules. The FLOPs of QGR depend only on the query length; we estimate its overhead using the average query length of 51 tokens across common benchmarks. For a fair comparison, we adopt the same compression setting as TokenPacker and LDP-V2, reducing the number of visual tokens from 576 to 144, which saves approximately 2.88T FLOPs in the LLM back-end. As shown in Table~\ref{tab-R2}, the additional projector overhead (at the scale of a few GFLOPs) is negligible compared to the compute saved by reducing LLM-side processing, confirming that QMoP’s efficiency gains are dominated by token reduction rather than projector computation.

%% file: sec/5_conclusion.tex
\section{Conclusion}
In this work, we present QMoP, a novel Query-Guided Mixture-of-Projector framework that adaptively integrates multiple visual token compression strategies via a learned routing mechanism, achieving state-of-the-art performance across a wide range of benchmarks.
To facilitate a deeper understanding of how compression affects multimodal model behavior, we introduce VTCBench, a dedicated diagnostic benchmark for visual token compression. 
Experiments demonstrate that QMoP not only surpasses existing compression baselines but also outperforms the uncompressed model in several scenarios, achieving a remarkable balance between efficiency and accuracy.  Additionally, we analyze the strengths and weaknesses of each compression strategy, providing insights and directions for future research on visual token compression.
% 在本文中，我们提出了一种查询引导的混合投影器（QMoP），这是一种用于多模态大型语言模型（MLLM）中视觉标记压缩的新颖而灵活的框架。QMoP 与查询引导路由器相结合，通过池化、剪枝和重采样三种互补压缩策略，自适应地压缩视觉标记。
% 为了评估低级视觉保真度，我们提出了 CPBench，这是一个专门评估全局理解、空间和数字意识、推理和常识、风格和情感感知以及局部细节保存的基准。
% 在 CPBench 和主流多模态基准上进行的实验表明，QMoP 不仅超越了现有的压缩基准，而且在多种情况下优于未压缩模型，在效率和准确性之间实现了显著的平衡。此外，我们总结分析了不同压缩策略的优缺点，对后续的visual token压缩提供了思路

%% file: supp_for_main.tex
\clearpage
\setcounter{section}{0}
\setcounter{subsection}{0}
\setcounter{subsubsection}{0}
\renewcommand{\theHsection}{supp.\arabic{section}}
\renewcommand{\theHsubsection}{supp.\arabic{section}.\arabic{subsection}}
\renewcommand{\theHsubsubsection}{supp.\arabic{section}.\arabic{subsection}.\arabic{subsubsection}}
\section*{Supplementary Material for QMoP}
\input{sec/X_suppl}

%% file: sec/X_suppl.tex
\section{Details of VTCBench}
This section provides a comprehensive description of VTCBench, including data sources, construction pipeline, annotation protocol, evaluation criteria, and dataset statistics.

\subsection{Data Sources for MTCBench}
To build a benchmark that captures diverse forms of visual degradation induced by token compression, we begin by pooling samples from six widely used multimodal datasets, covering a broad range of natural images and domain-specific scenarios. The selected datasets include:
\\
\textbf{MME.} Provides fine-grained visual perception and multi-skill evaluation, covering recognition, attribute understanding, and cross-modal grounding.
\\
\textbf{TextVQA} \cite{textvqa}.
Contains images rich in scene text, enabling evaluation of OCR-related reasoning, text grounding, and fine-grained local visual details.
\\
\textbf{GQA} \cite{gqa}.
Focuses on compositional reasoning, spatial understanding, and multi-step logical inference.
\\
\textbf{SEED-Image} \cite{seed}.
Offers diverse natural image scenarios designed for comprehensive vision–language evaluation with emphasis on semantic understanding.
\\
\textbf{MMStar} \cite{mmstar}.
Covers challenging real-world visual QA with complex semantics, cultural knowledge, and high-level commonsense reasoning.
\\
\textbf{SQA} \cite{sqa}.
Contains scientific diagrams and textbook-style images, emphasizing structured reasoning, visual abstraction, and domain-specific interpretation.

These six datasets collectively provide broad visual diversity (natural scenes, diagrams, text-heavy images, artistic variations) and rich reasoning types (spatial inference, commonsense reasoning, OCR, fine-grained recognition, scientific understanding), forming a solid foundation for constructing MTCBench.

\subsection{LLM-based Automatic Sample Filtering}
To ensure that each sample in MTCBench strictly aligns with one of the five evaluation dimensions, we employ an automatic filtering pipeline using Azure GPT-4.1 as the multimodal judge. The LLM is prompted to examine each (image, question, answer) triplet and determine whether it belongs to one of the predefined dimensions: Global Understanding (GU), Spatial and Counting (SC), Reasoning and Common Sense (RC), Style and Emotion (SE), or Local Details (LD). Samples that do not unambiguously match any dimension, or those containing invalid or low-quality annotations, are rejected. The exact prompt used for screening is provided below:
\newpage
\begin{tcolorbox}[float=t!,title={Prompt for Screening}]
\footnotesize
System content:

You are an expert multimodal evaluator. Your task is to determine whether a given
(image, question, answer) sample is suitable for inclusion in MTCBench, a benchmark
designed to evaluate visual token compression.

MTCBench contains exactly five evaluation dimensions:

1. Global Understanding (GU):
   Requires comprehension of the overall scene, major objects, high-level context,
   or broad semantic information.

2. Spatial and Counting (SC):
   Involves spatial relationships (e.g., left/right, in front of/behind) or counting
   the number of objects in the scene.

3. Reasoning and Common Sense (RC):
   Requires multi-step reasoning, logical inference, causal reasoning, or commonsense
   grounded in the visual content.

4. Style and Emotion (SE):
   Concerns artistic style, color composition, emotional expression, aesthetics, or
   affective interpretation conveyed by the image.

5. Local Details (LD):
   Requires fine-grained visual cues such as OCR (reading text), small objects,
   textures, or subtle details that are easily affected by token reduction.

Given an image and its QA pair, follow these steps:

Step 1:
  Determine whether the sample clearly belongs to one and only one of the five
  dimensions above. If it does not match any dimension, label it as "Reject".

Step 2:
  Validate that the question is grounded in the image and is not ambiguous,
  irrelevant, hallucinated, or extremely low quality. If any issue is found,
  label the sample as "Reject".

Step 3:
  If acceptable, assign exactly one of the dimensions:
  {GU, SC, RC, SE, LD}.

User content:

Please evaluate the following sample:

"Image": $<$Image$>$

"Question": $<$Question$>$

"Answer": $<$Answer$>$

Respond strictly in the following JSON format:

{
  "decision": "Accept" or "Reject",
  
  "dimension": "GU", "SC", "RC", "SE", "LD" or "null",
  
  "explanation": "short explanation (1-2 sentences)"
}

\end{tcolorbox}

\subsection{Human Verification and Balanced Dataset Assembly}
After automatic pre-classification, all remaining samples undergo human verification to ensure label accuracy and remove misclassified or ambiguous cases. Incorrect assignments are corrected, and invalid samples are removed.
Following verification, we assemble a balanced evaluation set of over 1,000 high-quality samples, ensuring that:
All five dimensions are equally represented.
Visual diversity within each dimension is preserved.
No single source dataset dominates the distribution.

\section{Mainstream Benchmark}
% 我们在GQA, VQA, TextVQA, MM-Vet, POPE, MME, MMB, Seed-Img, MMMU, MMStar这十个广泛使用的视觉理解基准上进行了实验。下面，我们将对这些基准进行详细描述。
We conduct experiments on ten widely used vision-understanding benchmarks: GQA \cite{gqa}, VQAv2 \cite{vqav2}, TextVQA \cite{textvqa}, MM-Vet \cite{mmvet}, POPE \cite{pope}, MME \cite{mme}, MMBench \cite{mmbench}, SEED-Image \cite{seed}, MMMU \cite{mmmu}, and MMStar \cite{mmstar}.
Below, we provide a detailed description of each benchmark.
\\
\textbf{GQA.} \cite{gqa}. GQA is a large-scale visual reasoning benchmark designed to evaluate a model’s ability to perform compositional, multi-step reasoning over real-world images. Questions are generated from scene graphs, ensuring precise grounding and explicit reasoning paths. Our method is evaluated on the subset of testdev balanced instructions, which includes 12,578 samples.
\\
\textbf{VQAv2.} \cite{vqav2}. VQAv2 is a widely used benchmark for open-ended visual question answering. Each question is paired with human-provided answers, reducing annotation bias and improving robustness. The benchmark measures a model’s ability to integrate visual recognition with commonsense reasoning across diverse question types, including counting, attribute recognition, and spatial understanding.
\\
\textbf{TextVQA.} \cite{textvqa}. TextVQA evaluates a model’s capability to read and reason over scene text in natural images. We evaluate the model’s performance on the test split, including 5,000 samples, where answering requires recognizing text via OCR and integrating it with visual context. This benchmark stresses the limitations of purely visual features and highlights the need for effective text–vision fusion.
\\
\textbf{MM-Vet.} \cite{mmvet}. MM-Vet is a recent benchmark designed to assess holistic multimodal reasoning. It comprises carefully curated questions that test capabilities such as object grounding, commonsense reasoning, mathematical understanding, instruction following, and multi-hop inference. Answers are evaluated using GPT-based scoring, making MM-Vet a challenging stress test for advanced multimodal models.
\\
\textbf{POPE.} \cite{pope}. POPE focuses on evaluating and mitigating object hallucination in multimodal models. It provides structured prompts requiring models to distinguish between present and absent objects, thereby measuring their tendency to generate hallucinated content. Performance on POPE reflects a model’s visual grounding fidelity.
\\
\textbf{MME.} \cite{mme}. MME is a large-scale multimodal evaluation suite comprising 14 distinct subtasks that target fine-grained perception and cognition abilities. It includes tests for object recognition, OCR, attribute understanding, spatial reasoning, and commonsense inference. MME provides a broad and detailed measurement of a model’s low-level and high-level perception accuracy.
\\
\textbf{MMBench.} \cite{mmbench}. MMBench is a comprehensive multiple-choice benchmark designed to measure general-purpose multimodal understanding across a wide range of dimensions, including world knowledge, logical reasoning, fine-grained visual recognition, and instruction following. Its diverse question set enables systematic comparison across different model families.
\\
\textbf{SEED-Image.} \cite{seed}. Seed-IMG is a multimodal benchmark designed to evaluate scene understanding, reasoning, and knowledge grounding from images using carefully curated question-answer pairs. The dataset emphasizes real-world scenarios, requiring models to combine perception with factual and commonsense reasoning.
\\
\textbf{MMMU.} \cite{mmmu}. MMMU targets expert-level reasoning, spanning more than 30 academic subjects, including medicine, chemistry, engineering, and economics. Questions often require solving multi-step problems by integrating diagrams, charts, and domain-specific knowledge.
\\
\textbf{MMStar.} \cite{mmstar}. MMStar focuses on fine-grained multimodal reasoning and visual comprehension across diverse scenarios. It includes tasks involving attribute recognition, spatial reasoning, multi-object interactions, commonsense inference, and open-ended grounding.

\begin{figure}[t!]
  \centering
  \includegraphics[width=0.8\linewidth]{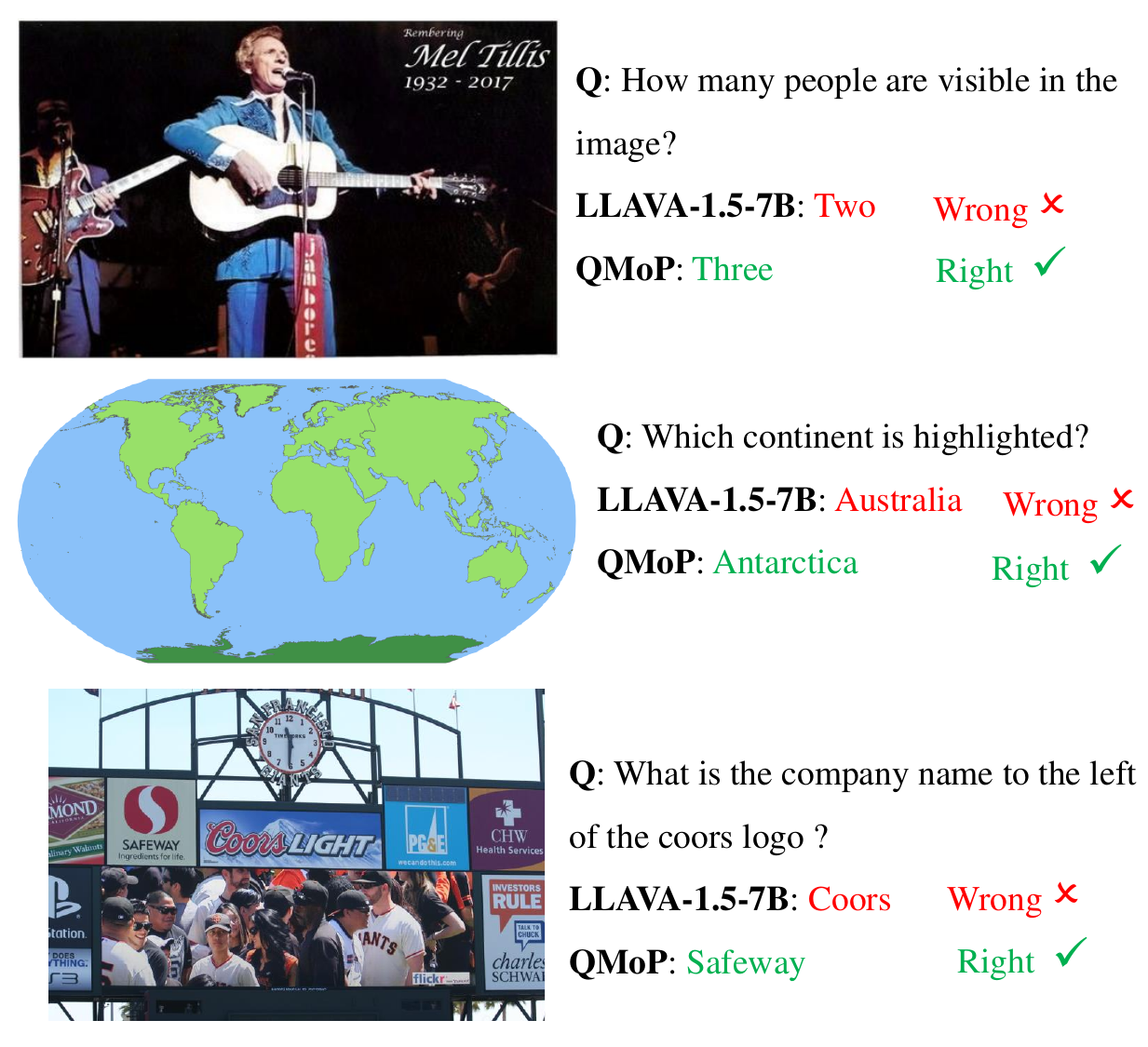}
  \caption{
  % LLaVA-1.5-7B 与我们提出的 QMoP 的定性比较。在多个场景中QMoP 提供了正确答案，而 LLaVA-1.5-7B 则失败。这些结果凸显了 QMoP 通过自适应标记压缩保留关键视觉信息的能力。
    Qualitative comparison between LLaVA-1.5-7B and our proposed QMoP. 
    }
  \label{fig-case}
\end{figure}
\section{Case Study}
To further illustrate the effectiveness of QMoP, we present a series of qualitative comparisons between our method and the baseline LLaVA-1.5-7B \cite{llava15}, as shown in Fig. \ref{fig-case}. In the first example, QMoP correctly counts three singers, including one partially hidden in shadow, whereas LLaVA detects only two, demonstrating QMoP’s ability to preserve occluded details and maintain numerical accuracy. In the second example, which requires interpreting a map’s orientation, QMoP integrates visual parsing with logical inference to answer directional queries accurately, while LLaVA’s response is irrelevant, highlighting QMoP’s superior reasoning in structured visual domains. In the final example, QMoP attends to subtle patterns and small object features that LLaVA overlooks, underscoring its sensitivity to fine-grained details. Across these scenarios, QMoP consistently delivers more accurate and context‑aware responses by retaining critical visual information, even outperforming the original uncompressed model in certain scenes.

% \label{sec:rationale}
% % 
% Having the supplementary compiled together with the main paper means that:
% % 
% \begin{itemize}
% \item The supplementary can back-reference sections of the main paper, for example, we can refer to \cref{sec:intro};
% \item The main paper can forward reference sub-sections within the supplementary explicitly (e.g. referring to a particular experiment); 
% \item When submitted to arXiv, the supplementary will already included at the end of the paper.
% \end{itemize}
% % 
% To split the supplementary pages from the main paper, you can use \href{https://support.apple.com/en-ca/guide/preview/prvw11793/mac#:~:text=Delete%20a%20page%20from%20a,or%20choose%20Edit%20%3E%20Delete).}{Preview (on macOS)}, \href{https://www.adobe.com/acrobat/how-to/delete-pages-from-pdf.html#:~:text=Choose%20%E2%80%9CTools%E2%80%9D%20%3E%20%E2%80%9COrganize,or%20pages%20from%20the%20file.}{Adobe Acrobat} (on all OSs), as well as \href{https://superuser.com/questions/517986/is-it-possible-to-delete-some-pages-of-a-pdf-document}{command line tools}.